\documentclass[conference]{IEEEtran}
\IEEEoverridecommandlockouts
\usepackage{cite}
\usepackage{amsmath,amssymb,amsfonts}
\usepackage{algorithmic}
\usepackage{graphicx}
\usepackage{textcomp}
\usepackage{flushend}
\usepackage[letterpaper, top=0.75in, bottom=1.1in,left=0.65in, right=0.68in]{geometry}

\begin{document}

\title{IoT Security*\\
{\footnotesize \textsuperscript{*}Note: Sub-titles are not captured in Xplore and
should not be used}
\thanks{Identify applicable funding agency here. If none, delete this.}
}

\title{Transmitter Identification and Protocol Categorization in Shared Spectrum via Multi-Task RF Classification at the Network Edge}

\author{\IEEEauthorblockN{Tariq~Abdul-Quddoos,~Tasnia Sharmin,~Xiangfang Li, Lijun Qian}
\IEEEauthorblockA{CREDIT Center and Department of Electrical and Computer Engineering  \\
Prairie View A\&M University, Texas A\&M University System  \\
Prairie View, TX 77446, USA \\
Email: \{tabdulquddoos, tsharmin, xili, liqian\}@pvamu.edu}
}


\maketitle

\begin{abstract}
As spectrum sharing becomes increasingly vital to meet rising wireless demands in the future, spectrum monitoring and transmitter identification are indispensable for enforcing spectrum usage policy, efficient spectrum utilization, and network security. This study proposed a robust framework for transmitter identification and protocol categorization via multi-task RF signal classification in shared spectrum environments, where the spectrum monitor will classify transmission protocols (e.g., 4G LTE, 5G-NR, IEEE 802.11a) operating within the same frequency bands, and identify different transmitting base stations, as well as their combinations. A Convolutional Neural Network (CNN) is designed to tackle critical challenges such as overlapping signal characteristics and environmental variability. The proposed method employs a multi-channel input strategy to extract meaningful signal features, achieving remarkable accuracy: $90$\%  for protocol classification, $100$\%  for transmitting base station classification, and $92$\% for joint classification tasks, utilizing RF data from the POWDER platform.
These results highlight the significant potential of the proposed method to enhance spectrum monitoring, management, and security in modern wireless networks. 
\end{abstract}


\section{Introduction}
In recent years, wireless services have become indispensable for people's everyday life and they rely on seamless spectrum access. From ground-based operations to air, sea, and space, wireless systems are fundamental to safeguarding national security and delivering essential public services that benefit society. Driven by continuous innovations in wireless technologies, the demand for spectrum access is growing at an unprecedented pace, underscoring its importance as a cornerstone of modern life and governance~\cite{NTIA}. However, expanding wireless network capabilities faces significant challenges due to the limited availability of spectrum. Spectrum serves as the foundation for bandwidth, and the industry's demand for more spectrum is critical to achieve higher data rates and support advanced use cases. Yet, the availability of unoccupied spectrum below 6 GHz is exceedingly scarce, intensifying the complexity of meeting these growing demands \cite{NI2019}. 

\begin{figure}
	 \centering
    	 \includegraphics[width=0.5\textwidth]{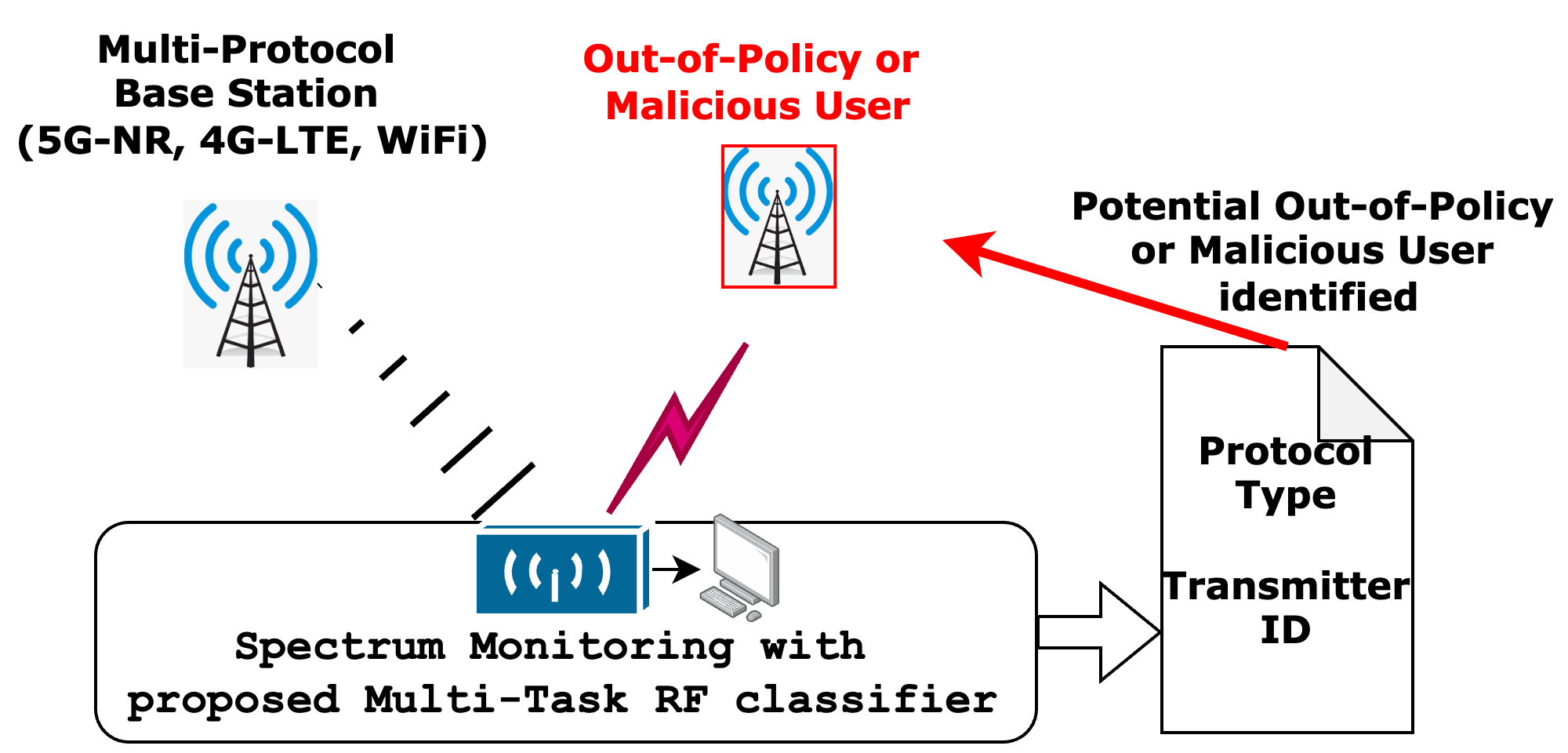}
      	\caption{Proposed Framework for transmitter identification and protocol categorization via multi-task RF signal classification in shared spectrum.}
    \label{fig:MotivationV2}
\end{figure}

Dynamic spectrum sharing is a promising strategy to address the increasing demands for wireless connectivity \cite{NTIA}. To further meet the increasing demand for spectrum, regulatory bodies such as European Commission and the Federal Communications Commission (FCC) in the United States have initiated study to assess the feasibility of unlicensed operations in the 6 GHz band. The FCC, through a Notice of Proposed Rule Making, sought public input on opening the $5.925–7.125$ GHz spectrum for unlicensed access in the United States~\cite{FCC2023, FCC2024}. Spectrum sharing enables harmonious operation among technologies such as Wi-Fi, private LTE, 5G New Radio Unlicensed (NR-U), Unlicensed LTE (LTE-U), License Assisted Access (LAA), MulteFire, and others, fostering efficient and equitable use of the available bandwidth \cite{girmay2023technology}.

In this evolving landscape, signal classification is pivotal for efficient spectrum management, interference mitigation, and more importantly, enhancing the security of wireless communication systems. While earlier research primarily focused on classifying signals based on their modulation type such as~\cite{ModulationRecognitionOshea2017}, 
the objective of this research is to develop a novel framework for RF signal classification capable of identifying and differentiating between multiple transmission protocols (4G LTE, 5G NR, IEEE 802.11a(WiFi)) operating within the same frequency bands, as well as distinguishing between transmitting base stations and their combinations, as highlighted in Fig.\ref{fig:MotivationV2}. 
This would allow spectrum monitor running the proposed scheme to identify potential out-of-policy or malicious users that transmit in frequency bands or time slots that they are not supposed to transmit.

The contributions of this study include
\begin{itemize}
    \item This study addresses critical challenges such as overlapping signal characteristics and environmental variations, ensuring robust and efficient classification for practical applications in spectrum management and security. 
    \item Unlike existing methods that focus on narrow dimensions or individual tasks, this research aims for joint classification by simultaneously analyzing protocols, base stations, and their combinations. It provides a robust solution to the complexities posed by overlapping signal features, particularly for closely related protocols like 4G LTE and 5G NR. 
    \item The proposed method employs a multi-channel input strategy to extract meaningful signal features, achieving remarkable accuracy: $90$\%  for protocol classification, $100$\%  for transmitting base station classification, and $92$\% for joint classification tasks, utilizing RF data from the POWDER platform.
    \item The proposed CNN based solution is light-weight and thus suitable for spectrum monitors at the network edge with limited resources.
\end{itemize}

The remainder of the paper is structured as follows: Section II reviews related works. Section III describes the dataset collection process and problem formulation. The proposed methodology is outlinied in Section IV. Section V presents experimental results and performance evaluations. Section VI concludes the study.






\section{Related Work}
Recent advancements in RF fingerprinting have explored the use of deep learning to address the limitations of traditional methods. In~\cite{zong2020rf}, the authors proposed a CNN-based approach that leverages the unique imperfections in RF transmitting circuits for device identification. By employing the Short-Time Fourier Transform (STFT), RF signals are transformed into time-frequency domain representations, which enhance feature extraction. Based on VGG-16, the study introduced an improved CNN model that demonstrated a good model which is able to identify transmitters effectively. Another study \cite{traynor2024physical} focuses on physical-layer security for IoT devices using RF fingerprints to provide fast and robust node authentication. The paper discusses various machine learning-based techniques such as DNNs, CNNs, and RNNs, along with non-ML approaches like statistical and information-theoretic methods. Notably, it proposes a novel method using Singular Value Decomposition (SVD) for efficient identification showcasing its suitability for secure and real-time IoT applications. In addition, authors in \cite{subray2021towards} explore the use of autoencoders to enhance spectrum sensing and classify signals, specifically distinguishing between LTE and Wi-Fi transmissions in shared radio spectrum environments. The research implements and evaluates three types of autoencoders—deep autoencoder, variational autoencoder, and LSTM autoencoder—based on their ability to classify signals using in-phase and quadrature (I/Q) data, amplitude, and phase features. Among the models, the deep autoencoder demonstrated the best precision, achieving up to 99.9\%, and was the fastest to train. 

The authors of \cite{girmay2023technology} tried to recognize technologies such as LTE, 5G NR, Wi-Fi, C-V2X PC5, and ITS-G5 in the 5.9 GHz Intelligent Transportation Systems (ITS) band. By utilizing CNN models trained with IQ sample Fast Fourier Transform (FFT) representations and a short Time Resolution Window (TRW) of 44 µs, the approach achieved over 97\% accuracy at a 20 Msps sampling rate. Also, paper \cite{bassey2019intrusion} utilize CNN is to extract device-specific features from RF traces, followed by dimensionality reduction using t-SNE and clustering with DBScan to identify unauthorized devices. Tested on RF data collected from six ZigBee devices, the training data does not have data of one device and the proposed they approch with that they were able to classify unauthorized device successfully. In~\cite{roy2019rfal}, the authors tried to use Generative Adversarial Networks (GANs) for RF transmitter identification and classification. The GAN framework consists of a generator that mimics trusted transmitters and a discriminator that identifies counterfeit signals, achieving 99.9\% accuracy in detecting adversaries. After eliminating rogue transmitters, the system classifies trusted transmitters using CNN, DNN, and RNN models, with RNNs achieving the highest accuracy of 97.85\%. Another study~\cite{POWDERRF}proposes a method for enhancing trust in Open RANs using RF fingerprinting on the POWDER PAWR platform. A CNN with a triplet loss function achieved 99.86\% accuracy in identifying base stations under varying channel conditions and across multiple waveforms such as WiFi, LTE, and 5G NR. Lastly, The authors in \cite{mohammad2023learning} used an innovative framework combines amplitude and phase spectrograms to identify wireless devices. By leveraging a CNN-LSTM architecture, the method extracts robust RF features and outperforms amplitude-only methods due to the additional information captured in the phase component. Tested on a dataset of LoRa devices, the framework employs machine learning classifiers and a triplet loss function, achieving high classification accuracy and better generalization for unseen devices.

To the best of our knowledge, previous research utilizing deep neural networks in RF fingerprinting has predominantly focused on specific tasks such as device identification through RF imperfections, technology recognition, or intrusion detection in wireless networks. In contrast, our research uniquely employs CNNs to perform combined joint classification by identifying signals based on base station, protocol, and their combination. 

\section{Dataset Collection and Problem Formulation}

\subsection{Dataset Collection}
\label{sec:dataset}
\begin{figure}
	 \centering
    	 \includegraphics[height=0.26\textheight]{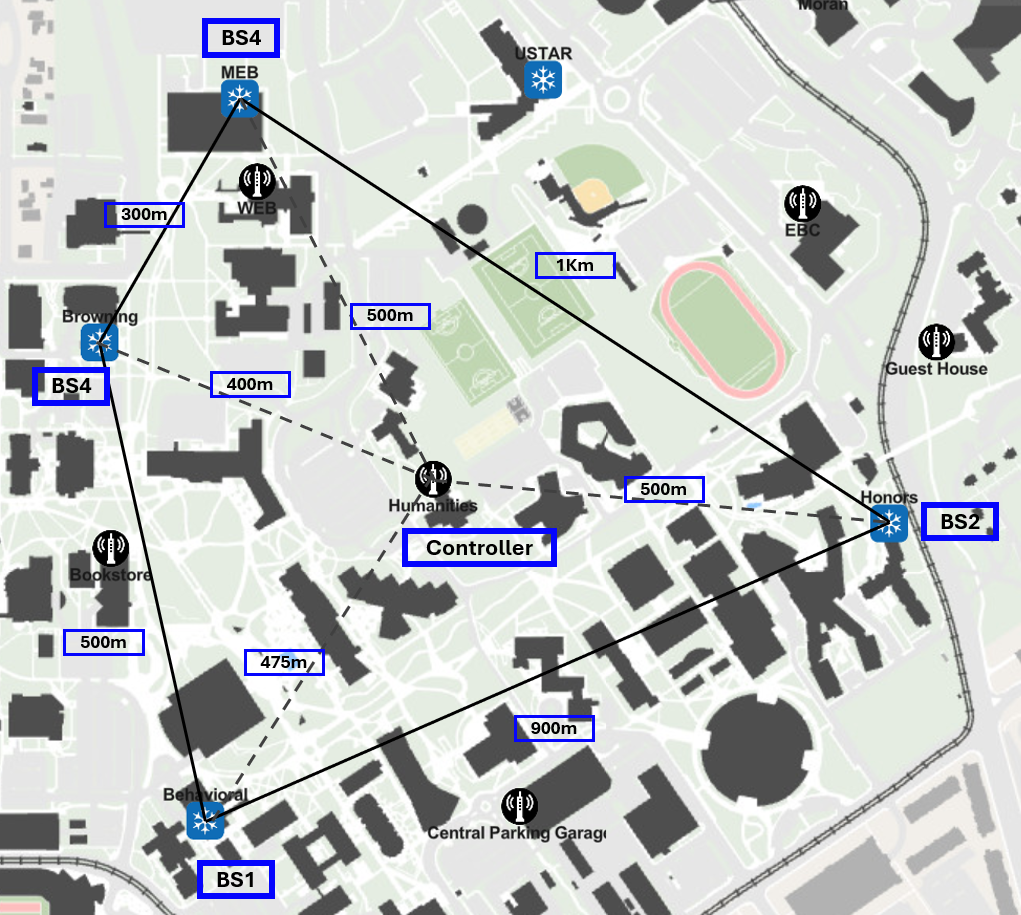}
      	\caption{POWDER Cite Map. Adapted from ~\cite{POWDERPlatform}.}
    \label{fig:POWDERMap}
\end{figure}

The RF data used in this work is a real-world dataset from the University of Utah's POWDER testbed, taken by a team from the Institute for Wireless Internet of Things at Northeastern University~\cite{POWDERRF}. The POWDER (Platform for Open Wireless Data-driven Experimental Research) is a city-scale software-defined platform for wireless research~\cite{POWDERPlatform}. The dataset consist of 120 signals each with 3 million IQ samples across 2 different days with 60 signals sampled per day. For a single day 5 sets of signals are taken for a particular transmitter base station and transmission protocol, with a total of 4 base stations and 3 protocols. The base stations used are MEB, Browning, Honors, and Behavioral as shown in figure.\ref{fig:POWDERMap}. The transmission protocols are IEEE 802.11a(WiFi), 4G LTE, and 5G NR. All signals have a center frequency of 2.685 GHz, with sampling rates of 5 MS/s for WiFi and 7.68 MS/s for 4G and 5G.

\subsection{Problem Formulation}
\label{sec:Problem}
In this work three problems are considered and formulated as classification problems.

The first is classifying the transmission protocol with classes being 4G LTE, 5G NR and IEEE 802.11a.
\begin{equation}
{\bf (P.1)}: \hspace{0.1in}
\hat{y}_{protcol} = f(x; \theta) \in \{ \text{4G LTE}, \text{5G}, \text{802.11a} \}
\label{problem1}
\end{equation}
The second is classifying the transmitting base station with classes being bes, browning, honors, and meb (names of the 4 base stations).
\begin{equation}
{\bf (P.2)}: \hspace{0.1in}
\hat{y}_{transmitter} = f(x; \theta) \in \{ \text{bes}, \text{browning}, \text{honors}, \text{meb}\}
\label{problem2}
\end{equation}
The third is classifying both the transmission protocol and transmitting base station jointly with a total of 12 classes, made up of combinations of the 3 protocols and 4 base stations.
{\bf (P.3)}: 
\begin{equation}
\begin{split}
\hat{y}_{protocol \cap trans.} &= f(x; \theta) \in \{ \text{4G LTE}, \text{5G}, \text{802.11a} \} \\
&\cap\ \{\text{bes}, \text{browning}, \text{honors}, \text{meb}\}
\end{split}
\label{problem3}
\end{equation}
\section{Proposed Methods}
\label{sec:method}

\subsection{Model Selection}
A convolutional neural network (CNN) is adopted as the classifier's model architecture. Past work has shown that CNNs work very well for extracting features from signals~\cite{girmay2023technology, POWDERRF, zong2020rf,bassey2019intrusion}. CNN's scale very well in terms of computational time in relation to computational resources as they are very parallelizable, as all kernels for the same layer can be done at the same time. This is good for signals as often they do not require heavy feature extractors. This combination of parallelization and lightweight modeling makes CNN's a good choice, especially for computing architectures that emphasize parallel operations like GPU's. The lightweight CNNs designed in this study also enable near-real-time RF signal classification at the edge where only limited computing resources are available.

\subsection{Model Architecture}
The CNN architecture in this work follows the structure of the CNN classifier from ~\cite{girmay2023technology}, with 1D Convolutions instead of 2D. The classifier takes a 1D sequence with 4 Channels, where the channels in this order are the signals real component, imaginary component, magnitude, and phase. The sequence is then passed through 3 1D convolution layers each with a kernel size of 9, stride of 1 and output channels sizes of 64, 32, and 16. Additionally batch normalization, max pooling with a stride and kernel size of 2, and dropout with a rate  of 0.5 are added to each layer. After the convolution layers the feature maps are flattened and passed through 2 feed-forward layers, the first making the flattened sequence into a length of 256 with a dropout rate of 0.3, and the second making the sequence the length of the number of classes. All layers have a ReLu activation function except the last where a Softmax function is applied to get a probability distribution.

\section{Results}
\label{sec:results}

\subsection{Experimental Setup}
For all three problems (P.1), (P.2), and (P.3), a total of $50000$ signals of length $1024$ are randomly sampled across the dataset with an uniform distribution across classes. There are $38000$ samples used for training, $2000$ samples are used for validation, and $10000$ samples are used for testing. The cross-entropy loss function is used. For the case of joint classifier, all combinations between protocols and base stations are treated as individual classes. The models are trained for 100 epochs with a batch size of 256 and learning rate of 0.001. The CNN based models come out to have a total of $522323$ parameters for the protocol classifier (3 classes), $522580$ parameters for the transmitter classifier (4 classes), and $524636$ parameters for the joint classifier (12 classes). 

The metrics applied are precision, recall, and F1-Score for  per-class evaluation and accuracy for overall results on a particular problem. Additionally, analysis is done using confusion matrix and t-SNE~\cite{TSNE} embeddings for more insights on models' performance. The training time with respect to validation performance is also analyzed as one of the goals of this work is to develop efficient solutions that suitable for resource-limited spectrum monitor at the network edge. 

\subsection{Results and Analysis}
\subsubsection{Protocol Classification}
\begin{figure}
	 \centering
    	 \includegraphics[width=0.4\textwidth]{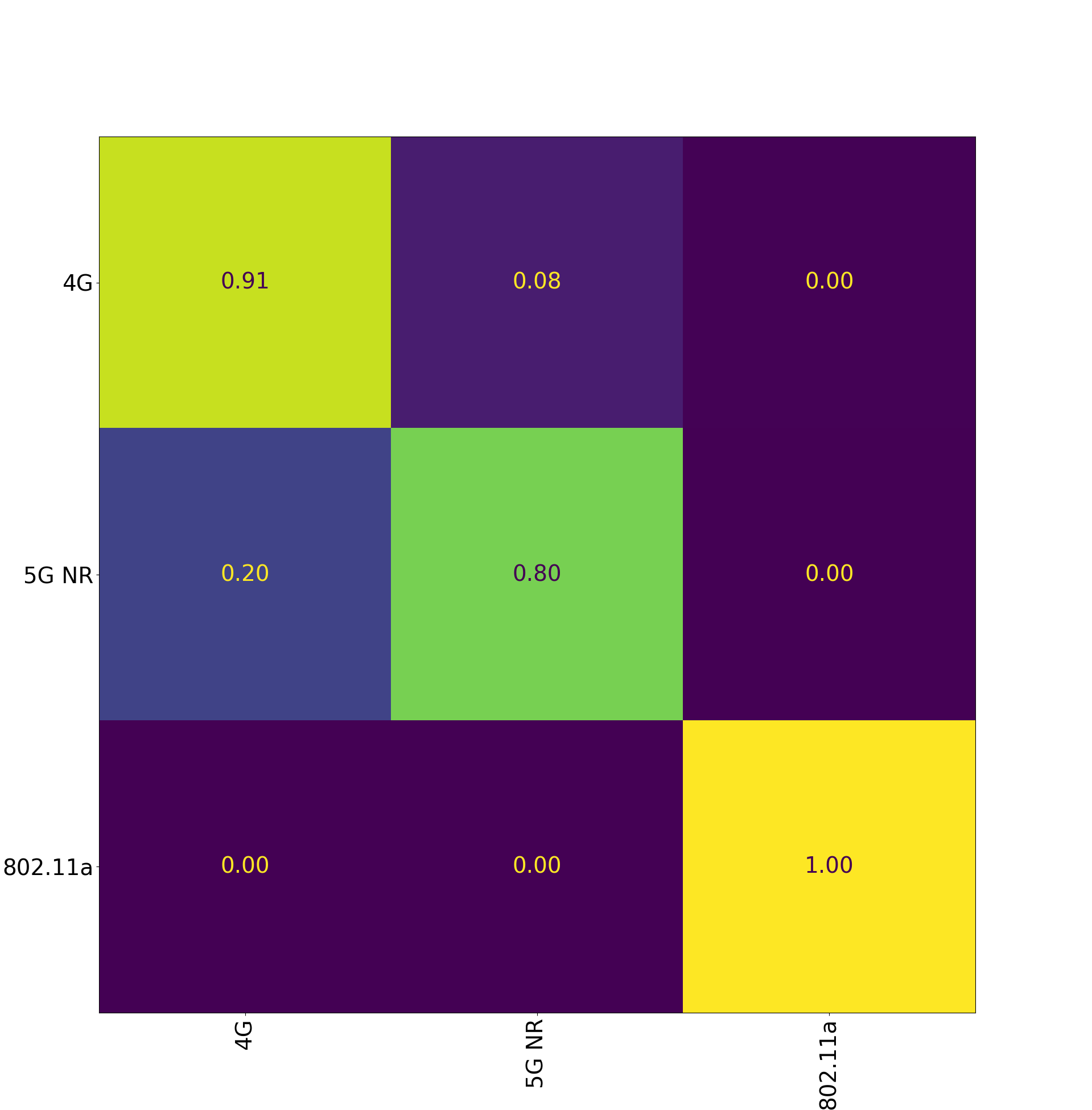}
      	\caption{Confusion Matrix of Protocol Classification}
    \label{fig:protcol_cm}
\end{figure}

\begin{figure}
	 \centering
    	 \includegraphics[width=0.5\textwidth]{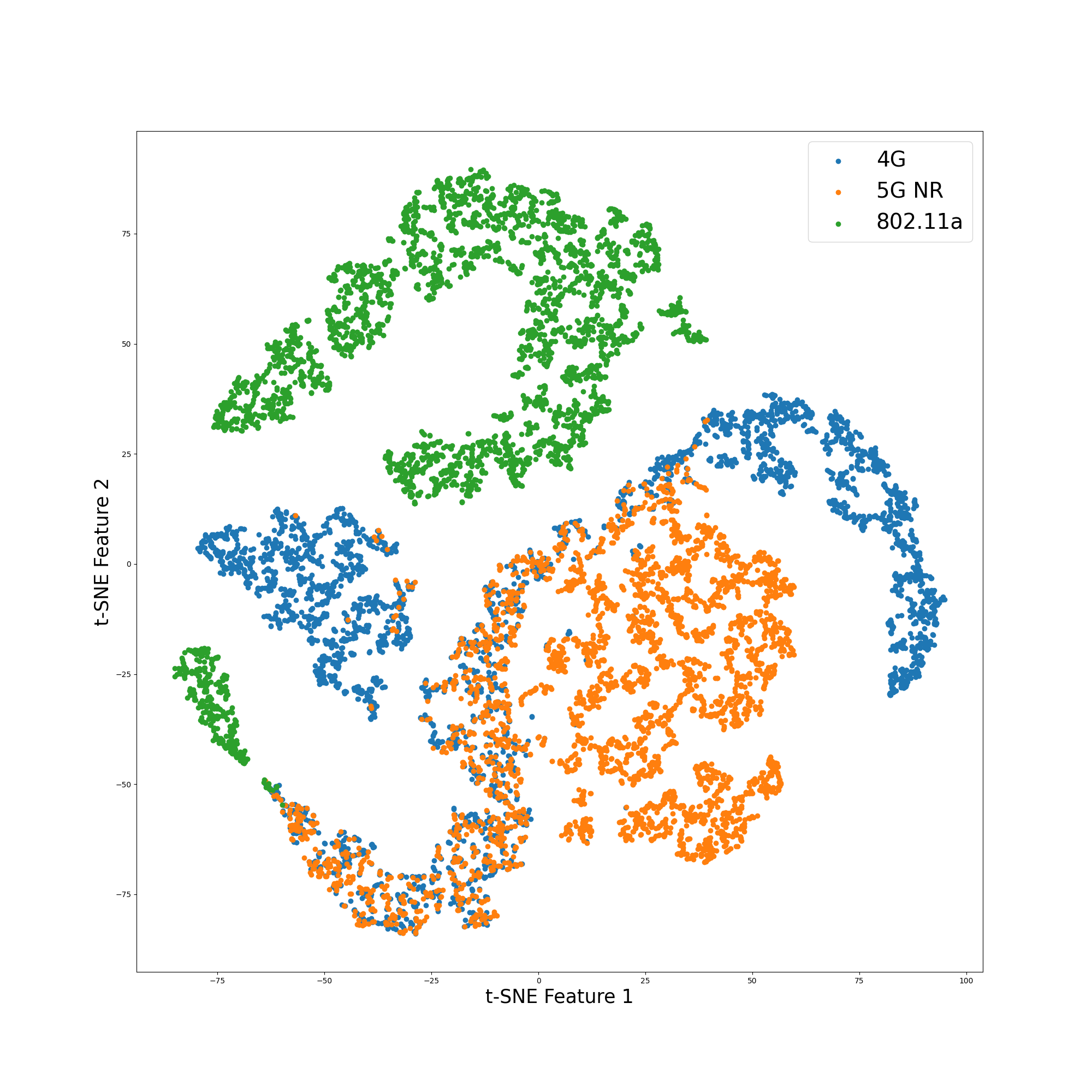}
      	\caption{Protocol T-SNE Embeddings}
    \label{fig:protocol_tsne}
\end{figure}

The results for classifying the protocols are shown in Table~\ref{table:protocol_results}, with an overall accuracy of $0.86$. The 802.11a class has a perfect performance with an F1-Score of 1.0, while 4G and 5G show to be classes contributing to the loss in performance having F1-Scores of 0.85 and 0.85. These observations are backed up by the confusion matrix shown in Fig.\ref{fig:protcol_cm} and the T-SNE embeddings shown in Fig.\ref{fig:protocol_tsne}, where there is clearly an overlapping of the 4G (blue) and 5G (orange) classes. 

\begin{table}[ht]
\centering
\begin{tabular}{|l|c|c|c|}
\hline
\textbf{Class} & \textbf{Precision} & \textbf{Recall} & \textbf{F1-Score} \\
\hline  
4G & 0.81 & 0.91 & 0.86 \\
5G NR & 0.91 & 0.80 & 0.85 \\
802.11a & 1.0 & 1.0 & 1.0 \\
\hline
\textbf{Accuracy} & \multicolumn{3}{c|}{0.90}\\
\hline
\end{tabular}
\caption{Protocol Classification Results}
\label{table:protocol_results}
\end{table}

\subsubsection{Transmitter Classification}
\begin{figure}
	 \centering
    	 \includegraphics[width=0.4\textwidth]{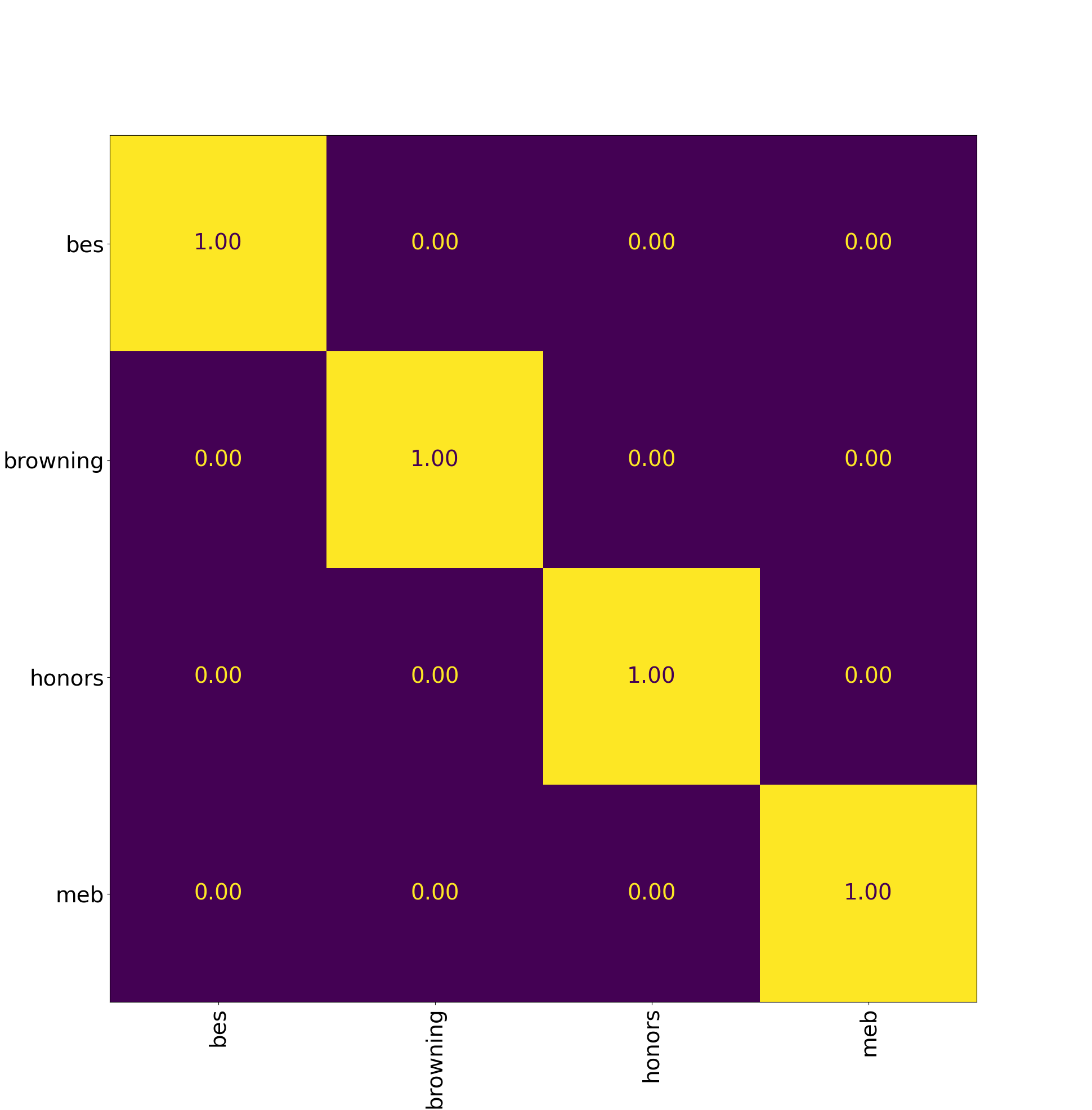}
      	\caption{Confusion Matrix of Transmitter  Classification}
    \label{fig:transmitter_cm}
\end{figure}

\begin{figure}
	 \centering
    	 \includegraphics[width=0.5\textwidth]{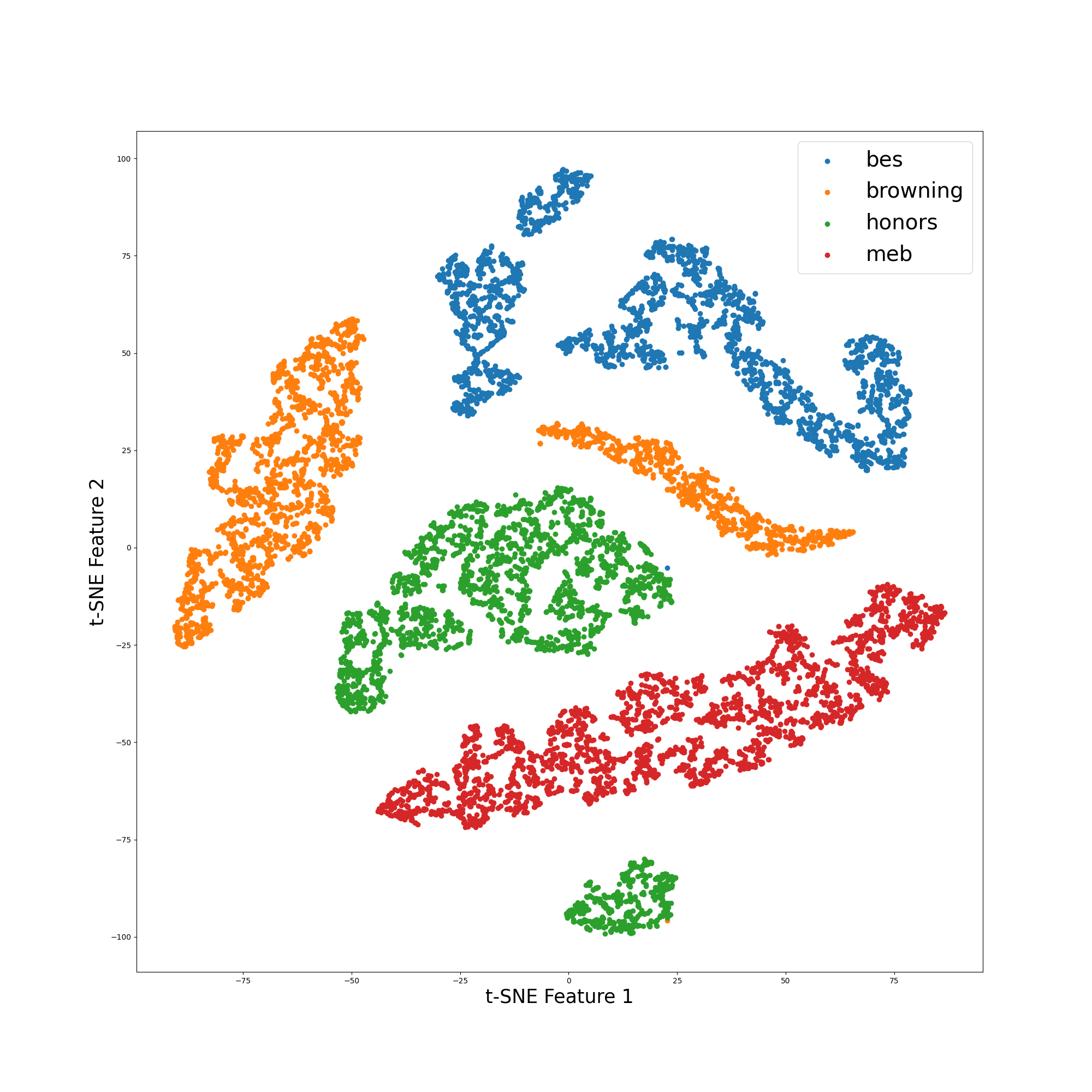}
      	\caption{Transmitter T-SNE Embeddings}
    \label{fig:transmitter_tsne}
\end{figure}

The results for classifying the transmitting base station are shown in Table~\ref{table:Transmitter_results}, with an overall accuracy of 1.0, with all classes have an F1-score of 1.0. The transmitter's confusion matrix is shown in Fig.\ref{fig:transmitter_cm}. It is observed that all the 4 classes representing the 4 transmitting base stations are non-overlapping as shown in the T-SNE embeddings in Fig.\ref{fig:transmitter_tsne}. This allows perfect classification results obtained for classifying the transmitting base stations. 

\begin{table}[ht]
\centering
\begin{tabular}{|l|c|c|c|}
\hline
\textbf{Class} & \textbf{Precision} & \textbf{Recall} & \textbf{F1-Score} \\
\hline  
bes & 1.0 & 1.0 & 1.0 \\
browning & 1.0 & 1.0 & 1.0 \\
honors & 1.0 & 1.0 & 1.0 \\
meb & 1.0 & 1.0 & 1.0 \\
\hline
\textbf{Accuracy} & \multicolumn{3}{c|}{1.0}\\
\hline
\end{tabular}
\caption{Transmitter Classification Performance}
\label{table:Transmitter_results}
\end{table}

\subsubsection{Joint Protocol and Transmitter Classification}
\begin{figure}
	 \centering
    	 \includegraphics[width=0.4\textwidth]{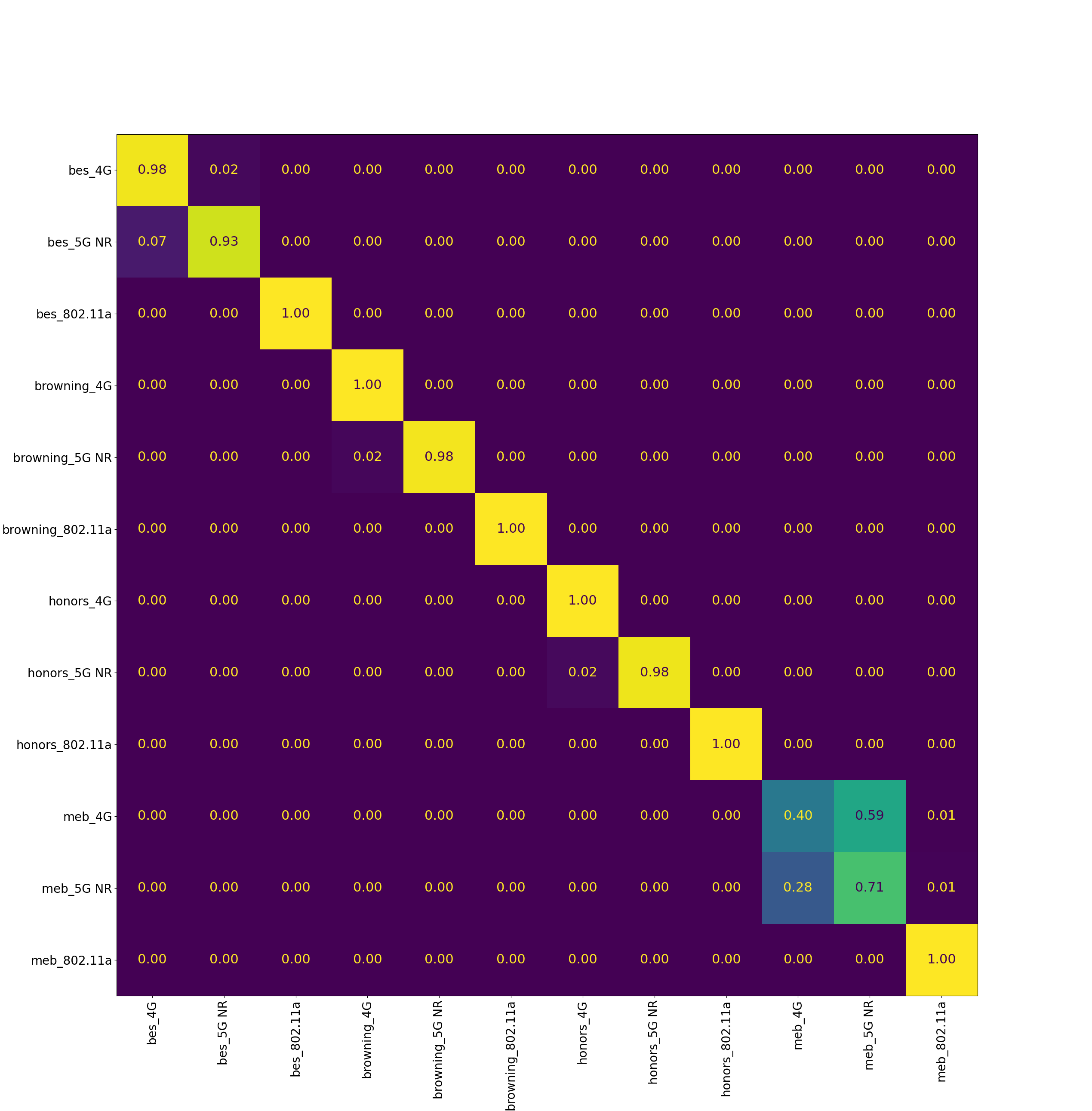}
      	\caption{Joint Confusion Matrix}
    \label{fig:joint_cm}
\end{figure}

\begin{figure}
	 \centering
    	 \includegraphics[width=0.5\textwidth]{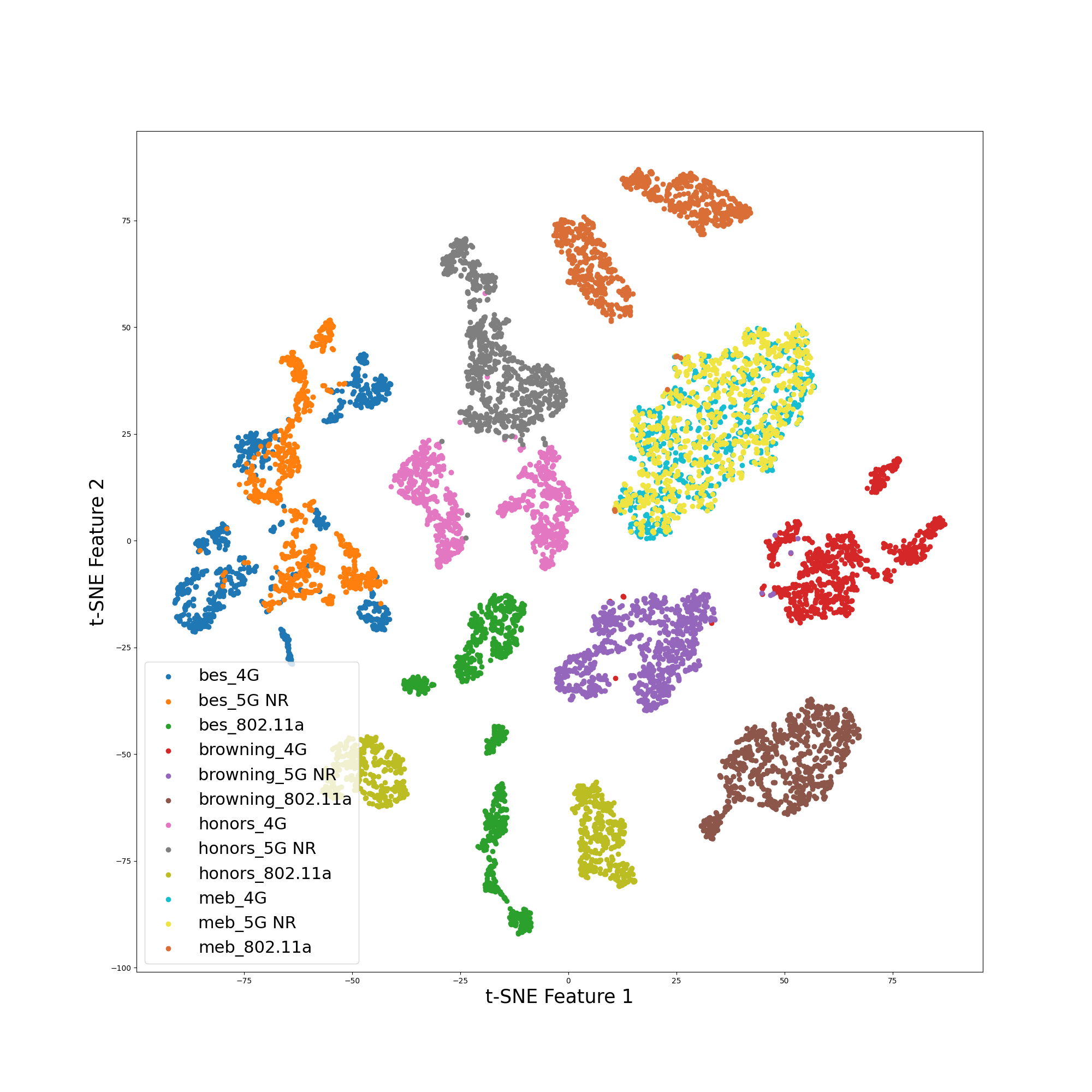}
      	\caption{Joint T-SNE Embeddings}
    \label{fig:joint_tsne}
\end{figure}

The results for classifying the protocol and transmitting base station jointly are shown in Table~\ref{table:Joint_results}, with an overall accuracy of 0.92. The primary classes contributing to the loss in performance are med\_4G and meb\_5G, with F1-scores of 0.47 and 0.62. The confusion matrix is given in Fig.\ref{fig:joint_cm}. The T-SNE embeddings given in Fig.\ref{fig:joint_tsne} show a complete overlap between the meb 4G (light blue) and 5G NR (yellow) classes, as well as some overlap between the bes 4G (blue) and bes 5G NR (orange) classes, provide a visual explanation of the results obtained. 
\begin{table}[ht]
\centering
\begin{tabular}{|l|c|c|c|}
\hline
\textbf{Class} & \textbf{Precision} & \textbf{Recall} & \textbf{F1-Score} \\
\hline  
bes\_4G & 0.93 & 0.97 & 0.96 \\
bes\_5G NR & 0.97 & 0.93 & 0.95 \\
bes\_802.11a & 1.0 & 1.0 & 1.0 \\
browning\_4G & 0.98 & 0.99 & 0.99 \\
browning\_5G NR & 0.99 & 0.99 & 0.99 \\
browning\_802.11a & 1.0 & 1.0 & 1.0 \\
honors\_4G & 0.97 & 0.99 & 0.98 \\
honors\_5G NR & 0.99 & 0.97 & 0.98 \\
honors\_802.11a & 1.0 & 1.0 & 1.0 \\
meb\_4G & 0.56 & 0.40 & 0.47 \\
meb\_5G NR & 0.56 & 0.71 & 0.62 \\
meb\_802.11a & 0.99 & 1.0 & 0.99 \\
\hline
\textbf{Accuracy} & \multicolumn{3}{c|}{0.92}\\
\hline
\end{tabular}
\caption{Joint Classification Performance}
\label{table:Joint_results}
\end{table}

\subsection{Discussions}
The results show that the proposed CNN classifier can distinguish transmitting base stations well but has some difficulty classifying protocols as it does not distinguish 4G and 5G as well. The joint classification results show that this is not necessarily true for all transmitters but primarily for base station meb and slightly for base station bes. These transmitters can be inferred to be the contributing factors to the loss in performance for classifying the protocols.

\begin{figure}
	 \centering
    	 \includegraphics[width=0.5\textwidth]{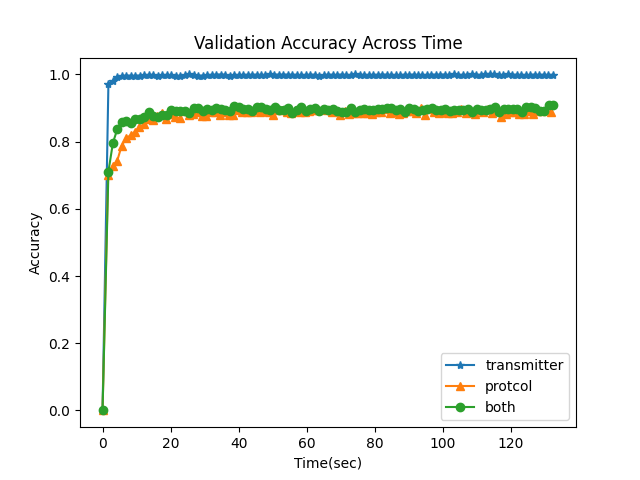}
      	\caption{Validation Results Across Time During Training}
    \label{fig:val_time}
\end{figure}

The computationally efficiency in terms of training is shown in Fig.\ref{fig:val_time}. The model can achieve around almost 1.0 accuracy for classifying the transmitting base stations in approximately $10$ seconds, and around 0.9 accuracy in approximately $20$ seconds for the protocol and joint classification. 

\section{Conclusions}
\label{sec:conclusions}
This research presents a comprehensive approach to RF signal classification, tackling critical challenges such as overlapping signal characteristics and environmental variability. By leveraging CNNs and RF data collected from the POWDER platform, the proposed framework achieves impressive accuracy across multiple classification tasks, including protocols (4G LTE, 5G NR, IEEE 802.11a), base stations, and their combinations. The proposed model employs a multi-channel input strategy to extract robust signal features, ensuring high accuracy and efficiency. It achieves 90\% accuracy in protocol classification, 100\% in base station classification, and 92\% in joint classification tasks. The CNN based classifiers also act as efficient modeling solutions, as they can achieve excellent performance in very short time (seconds) that make them promising solution for spectrum monitoring at the network edge with limited resources. 

As the world moves towards spectrum sharing to address growing wireless demands, transmitter identification and protocol categorization will become indispensable for efficient spectrum monitoring and management, and discovering out-of-policy or malicious users for robust network security. This study proposed light-weight CNN based classifiers by integrating protocol, base station, and joint classifications, a novel approach not previously explored in this context. The findings of this paper will make it a valuable tool for monitoring and managing the increasing complexities of wireless systems in shared spectrum, particularly as we transition into the 6G era.
Future research will focus on enhancing the scalability of the proposed approach, and evaluating its adaptability under rapidly changing spectrum conditions.

\section{Acknowledgments}
\label{acknowledgement}
This research work is supported by the U.S. Army Research Office (ARO) under grant number W911NF-23-1-0214 and the U.S. National Science Foundation (NSF) under award number 2302469. The views and conclusions contained in this document are those of the authors and
should not be interpreted as representing the official policies, either expressed or implied, of the ARO, NSF, or the U.S. Government. The U.S. Government is authorized to reproduce and distribute reprints for Government purposes notwithstanding any copyright notation herein.

\bibliographystyle{IEEEtran}
\bibliography{IOTSecurityJeff}

\end{document}